\documentclass{article}
\usepackage{graphicx} 
\usepackage{authblk}  
\usepackage{float}    
\usepackage{multirow} 
\usepackage{caption} 

\title{ Zero-shot Shape Classification of Nanoparticles in SEM Images using Vision Foundation Models }

\author[1]{Freida Barnatan}
\author[1]{Emunah Goldstein}
\author[1]{Einav Kalimian}
\author[1]{Orchen Madar}
\author[2]{Avi Huri}
\author[2]{David Zitoun}
\author[2,3]{Ya'akov Mandelbaum}
\author[1]{Moshe Amitay}

\affil[1]{Department of Bioinformatics, Jerusalem College of Technology\\
Jerusalem, Israel 91160\\
\texttt{mamitay@g.jct.ac.il}}

\affil[2]{Bina (Bar-Ilan Institute of Nanotechnology and Advanced Material), Bar-Ilan University\\
Ramat Gan, Israel 5290002}

\affil[3]{Department of Electrical Engineering / Department of Electrooptical Engineering and Applied Physics, Lev AC - Jerusalem College of Technology\\
Jerusalem, Israel 91160}

\date{May 2025}

\begin{document}

\maketitle

\section{Abstract}

 Accurate and efficient characterization of nanoparticle morphology in Scanning Electron Microscopy (SEM) images is critical for ensuring product quality in nanomaterial synthesis and accelerating development. However, conventional deep learning methods for shape classification require extensive labeled datasets and computationally demanding training, limiting their accessibility to the typical nanoparticle practitioner in research and industrial settings. In this study, we introduce a zero-shot classification pipeline that leverages two vision foundation models: the Segment Anything Model (SAM) for object segmentation and DINOv2 for feature embedding. By combining these models with a lightweight classifier, we achieve high-precision shape classification across three morphologically diverse nanoparticle datasets—without the need for extensive parameter fine-tuning. Our methodology outperforms a fine-tuned YOLOv11 and ChatGPT o4-mini-high baselines, demonstrating robustness to small datasets, subtle morphological variations, and domain shifts from natural to scientific imaging. Quantitave clustering metrics on PCA plots of the DINOv2 features are discussed as a means of assessing the progress of the chemical synthesis. This work highlights the potential of foundation models to advance automated microscopy image analysis, offering an alternative to traditional deep learning pipelines in nanoparticle research which is both more efficient and more accessible to the user. 

 Keywords: nanoparticle, synthesis, nanoscience, chemistry,  SEM, image, analysis, recognition, classification, DINO, SAM, zero-shot, deep learning AI

\section{Introduction}

Scanning Electron Microscopy (SEM) images provide detailed information about nanoparticles morphology such as shape and size.  These parameters directly influence the physical, chemical, and functional properties of nanoparticles, making their measurement vital for reliable nanoparticle manufacturing in fields ranging from drug delivery and diagnostics to catalysis and materials engineering.  Quantifying shape and size distributions plays also a critical role in the detection of potential contaminants and verification of batch-to-batch consistency in nanoparticle manufacturing.\cite{vladar2020characterization}

\begin{figure}[H]
    \centering
    \includegraphics[width=0.5\linewidth]{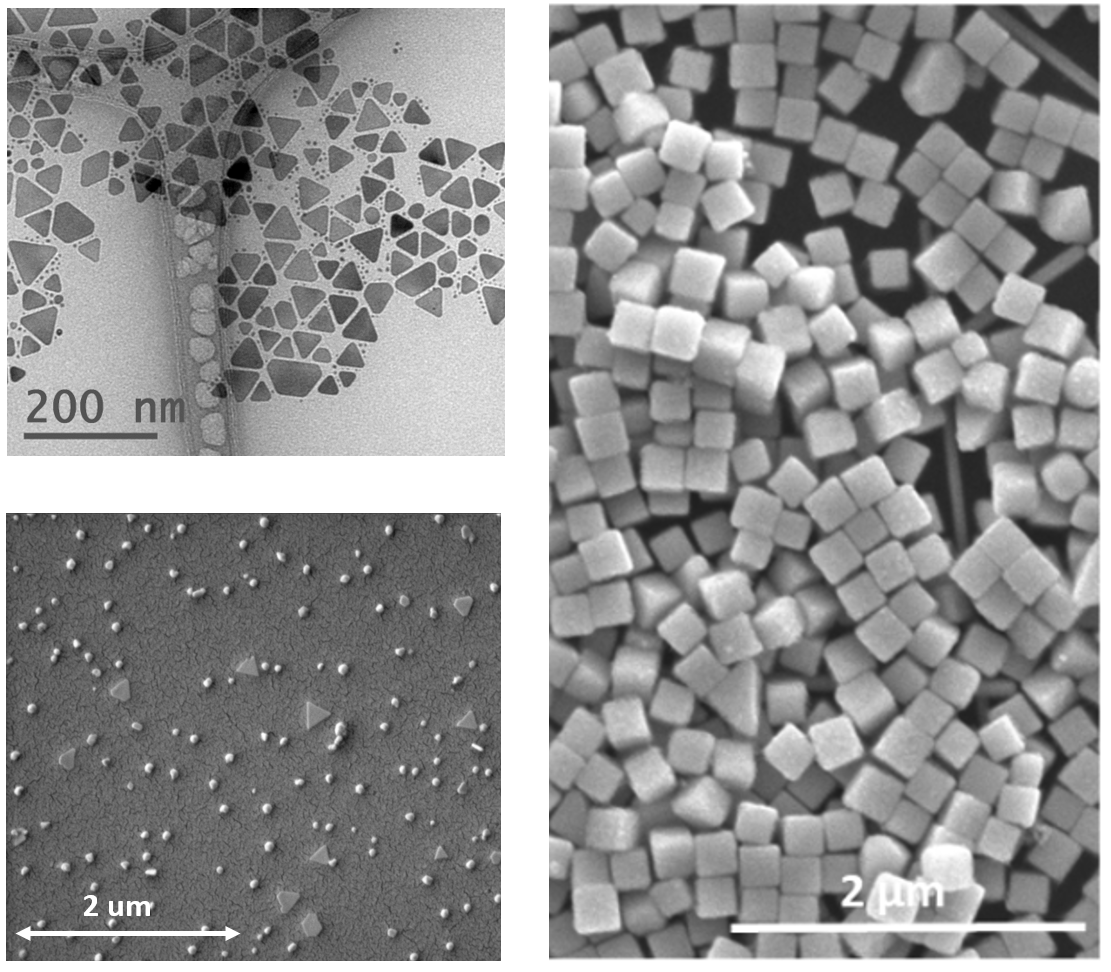}
    \caption{SEM images of nanoparticles of various morphologies: clockwise from lower left: spheres ('quantum dots'), triangles and cubes.}
    \label{fig:nanoparticles_training}
\end{figure}

Figure~\ref{fig:nanoparticles_training} displays SEM images of nanoparticles for a selection of different shapes - spheres, triangles and cubes.

However, extraction of nanoparticle batch properties from SEM images is not a trivial undertaking. Typically a batch numbers tens to hundreds of sample particles, so that simply assessing number can be time-consuming.  Determining the size requires detailed measurements by hand for each sample. Thus, the chemist will typically make do with a representative sampling, sufficient to estimate the average size and dimensions. An example demonstrating this can be viewed  in Figure~\ref{fig:manual-characterization} in the appendix.
A comprehensive extraction of the entire distribution remains a truly daunting task requiring hours of meticulous effort.
Last,  insofar as shape is concerned - for the lone chemist an impression made at-a-glance may suffice to decide on the quality of the product. Yet a determination of a quantitative metric of the morphology  of the batch -  sharpness of corners, eccentricity, symmetry etc. - essentially lies beyond the reach of a researcher using simple methods. Clearly then, characterization of nanoparticle products using manual analysis of SEM images is not scalable for high-throughput platforms.

Moreover automated synthesis platforms, which can produce nanoparticles with high throughput and reproducibility, have become increasingly prevalent.\cite{zhao2023robotic,zaki2025selfdriving,dembski2023robot,diao2024leveraging} These systems require robust downstream quality control to ensure that the resulting particles meet strict morphological and functional specifications. 
 We envision meeting the challenge by integrating computerized SEM image analysis into a feedback control loop, in order to enable intelligent tuning of synthesis parameters based on the outcomes  - number, size, morphology - of each batch. This loop could operate in either real-time or batch-wise modes, depending on the imaging and analysis throughput.\cite{Angello2022SuzukiMiyaura} For example, if SEM-based classification consistently detects that particles are larger than the target size, this may indicate issues such as insufficient nucleation or suboptimal reaction temperatures.  
 The framework of generative artificial intelligence (AI) is thus extended to include the synthesis: the detailed  time-log of the synthesis conditions - temperature, input flow rates of reactants mixing rates, in-situ spectrometric measurements etc. paired with the product characterization serves as the training data. This is then used by the system logic to create or modify the synthesis procedure and generate the control inputs for optimal operation thereafter.
 Such a feedback mechanism transforms the synthesis process from a step-by-step, trial-and-error workflow into a data-driven, self-correcting system which is particularly valuable in industrial-scale nanoparticle production.\cite{Taylor2023ReactionOptimization}

While many previous studies have employed traditional deep learning approaches such as object detection and segmentation networks to detect and classify nanoparticles in SEM images,\cite{Ragone2023DeepLearningMicroscopy} these methods present practical challenges. Deep learning architectures like YOLO\cite{lopeznanoparticle,redmon2016yolo} for object detection, Mask R-CNN\cite{monchot2021deep,he2017mask} for instance segmentation, and U-Net\cite{ruhle2021workflow,Bals2023NanoparticleSEM,ronneberger2015unet} for semantic segmentation all require task-specific fine-tuning, which involves preparing large training datasets, performing extensive manual labeling and  training heavy deep learning architectures with many layers and millions of parameters — often demanding significant computational resources and time investment \cite{chen2024advancing}. In addition, training deep learning typically demands substantial experience and careful tuning.  Practitioners must make informed decisions about architecture design, regularization techniques, learning rate schedules, and data augmentation strategies to achieve satisfactory performance .
This places these methods large out of reach for most in the chemisry and nanoparticle synthesis community, requiring the addition of one more dedicated professionals to the staff of any project.

In this study,  a methodology is presented that utilizes zero-shot (i.e: without training) foundation models for analyzing nanoparticle shapes in SEM images.\cite{zhou2024foundation} Foundation models are large, general-purpose deep learning models trained on massive and diverse datasets such as billions of images so they can learn broad, reusable patterns. Once trained, they can be applied to many different tasks (like classification, detection, or segmentation) with no extra training, often achieving high performance even in unfamiliar domains. Unlike traditional deep learning models like Convolutional Neural Networks (CNNs)\cite{Zhao2024CNNReview} or early Vision Transformers (ViTs)\cite{Han2023ViTSurvey}, which are often pretrained for specific domains or tasks (e.g., ImageNet for natural images)\cite{Deng2009ImageNet} and then fine-tuned through transfer learning,\cite{Panda2024TransferLearningSurvey} foundation models are trained on massive, diverse, and uncurated datasets across many domains, with the goal of learning general-purpose representations. While transfer learning with CNNs or ViTs typically requires task-specific fine-tuning on new labeled data, foundation models are often be applied to new problems without retraining.
In essence, foundation models differ in their pre-training scale, training objective, and versatility making them suitable as plug-and-play tools across a wide range of tasks. 

Specifically, we utilize two state-of-the-art models developed by Meta AI: the Segment Anything Model (SAM)\cite{kirillov2023segment} for segmentation and DINOv2\cite{oquab2023dinov2} \cite{doron2023unbiased} for feature extraction that subsequently used to train a lightweight, task-specific classifier—such as logistic regression—with minimal learnable parameters. This approach enables effective nanoparticle classification without the need to train or fine-tune a deep neural network, significantly reducing computational demands while preserving high performance. 
SAM is a powerful vision foundation model capable of segmenting arbitrary objects in images without prior exposure, thanks to its intensive pretraining on over a billion masks. The second model: DINOv2 is a powerful vision foundation model that generates rich image representation vectors, also known as image embeddings. DINOv2 excels at capturing abstract morphological patterns, texture, and spatial detail. These features are  also robust to variations in object position and orientation. By combining SAM for segmentation and DINOv2 for feature extraction, and training only a lightweight classifier with minimal learnable parameters, our approach enables robust classification of nanoparticle shapes. This methodology is particularly well-suited for microscopy applications, as it eliminates the need for large annotated datasets, extensive manual labeling, and GPU-intensive model training. As a result, automatic classification of nanoparticle shapes becomes more robust and accessible compared to conventional deep learning pipelines. 

In this work the methodology described is demonstrated on SEM images of the nanoparticle types sampled in Figure~\ref{fig:nanoparticles_training}. To the best of our knowledge, this is one of the first applications of DINOv2 for  nanoparticles classification. This domain fundamentally different from the natural image datasets on which the model was trained. Unlike typical image classification tasks, SEM images contain grayscale geometric variations of shapes, often with subtle morphological differences. These cut-out images of individual shapes are also significantly smaller in size compared to the images used during DINOv2's pretraining. Despite this extreme domain and scale shift, we explored the model’s ability to extract representations features.  

To benchmark the classification performances of this methodology,  two comparisons were made. First the ubiquitous ChatGPT - model 'o4-mini-high' a model optimized for visual reasoning  -  was used with a 'zero-shot' methodology to serve as base-line. Next the results were  evaluated against YOLOv11—a widely used deep learning model for object detection and classification - after fine-tuning on a training set.\cite{Khanam2024YOLOv11} 
This study delivers a practical and effective methodology for nanoparticle detection and shape classification in SEM images, enabling chemists to perform accurate analyses without deep learning expertise or extensive data. By combining foundation models in a zero-shot framework, the method outperforms both ChatGPT o4-mini-high, and a fine-tuned YOLOv11 baselines demonstrating its reliability and accessibility for real-world applications.

\section{Methods}

\textbf{1. Nanoparticle Datasets }

We used three types of nanoparticles  datasets:
The first dataset contains primarily cubic nanoparticles, along with a contamination of pyramid-shaped particles. The image presents an unordered mixture of these morphologies, posing a challenge for accurate classification.

The second dataset contains 2D triangular-shaped nanoparticles, and was used to assess the ability of the pipeline to capture and classify geometric features related to angularity and edge definition. Specifically, we aimed to evaluate the model’s performance in identifying and distinguishing particles based on their degree of triangularity. 

The third dataset, consists of a population of spherical (dot-like) nanoparticles intermixed with a smaller number of contaminating particles. The primary challenge in this case lies in the extremely small dimensions of the particles (around 30X30 pixels).

Samples from the three datasets are shown in Figure~\ref{fig:nanoparticles_training} and the quantities of training, validation, and test samples for each class are detailed in Table 1, with particular emphasis on the limited number of training samples, highlighting the challenge of learning with scarce training data. 

\begin{table}[ht]
\centering
\caption{Number of annotated particles per class used for train, validation and test.}
\begin{tabular}{|c|p{3.5cm}|c|c|c|}
\hline
\textbf{Dataset} & \textbf{Class}            & \textbf{Train} & \textbf{Validation} & \textbf{Test} \\
\hline
1 & Cubes               & 10 & 73& 176\\
 & Pyramids            & 10 & 58& 12 \\
2 & Triangles           & 7  & 6  & 83\\
 & Truncated triangles & 5  & 6  & 11 \\
 & Circles             & 6  & 7  & 11 \\
3 & Dots                & 6  & 9  & 111\\
 & Non-dots            & 6  & 8  & 7  \\
\hline
\end{tabular}
\label{tab:dataset_splits}
\end{table}

\textbf{2. Segmentation and Cropping Using SAM}
 To segment and extract individual nanoparticles from SEM images, we employed the Segment Anything Model1 (SAM1) with its zero-shot mask generation capability. Although SAM2 was recently developed we prefered to apply SAM1 since Sengupta \textit{et. al.} \cite{sengupta2025sam} study showed that SAM2 performs worse than SAM1 on low-contrast imaging modalities. Since SEM images of nanoparticles often exhibit similar challenges—such as low contrast and fuzzy edges—SAM2 is likely to underperform in this context, making SAM1 a more suitable choice. 
Usind SAM1  a binary mask was output for every detected particle. To ensure that the segmentation masks captured meaningful structures rather than noise, only the confident and stable masks were retained using thresholds for predicted IoU ($\geq 0.95$), stability score ($\geq 0.95$), and minimum region area ($\geq 500$ pixels). We then cropped each image to the bounding box defined by the upper-left and bottom-right pixels of the segmentation mask, ensuring the full extent of the particle was enclosed (See Figure~\ref{fig:binary_masks_thumbnails}.)
 
\textbf{3. Preparation of train, validation and test sets for each of the datasets}
 Cropped nanoparticle images were randomly selected for training and validation and then manually annotated with the corresponding class label. In fact for the triangles the original binary masks produced by SAM1 were input to the classifier for both training/validation and testing, as this reduces noise and was found to improve performance.  It should be noted that images for the cubes and triangles validation dataset were generated synthetically using DALL·E of OpenAI\cite{Ramesh2021ZeroShot} due to unavailability of data. A sample image is displayed in Figure~\ref{fig:synthetic_generated_validation} of the appendix.
For the test set, we labeled all the segmented particles in the entire SEM image used for the test. Table 1 summarizes the number of annotated particles per class used for training, validation, and testing.  

\textbf{4. DINOv2 Feature Extraction and Classifier}
We extracted feature embeddings from the cropped images in the training, validation, and test sets using the pretrained DINOv2 ViT-B/14 model. Prior to embedding extraction, various resizing and padding transformations were applied to center and standardize the small input image dimensions to 224×224 pixels. To identify the most effective preprocessing strategy prior to embedding with DINOv2, we evaluated several candidate pipelines based on their ability to maximize class separation. For each preprocessing method, DINOv2 embeddings were generated, and the mean feature vector (centroid) was computed for each class. We then calculated the average euclidean distance between all pairs of class centroids. This average distance served as a quantitative measure of how well the preprocessing preserved or enhanced inter-class discriminability. The preprocessing approach that yielded the highest average distance between class centroids was selected as optimal, as it indicated improved separation among classes. 

For the classification step based on DINOv2 embeddings, we employed two approaches were employed. In the first, we performed hyperparameter optimization on the validation set using grid search to select the best-performing model. We tried multiple classifiers: logistic regression, support vector machines, random forests, and Naive Bayes with multiple preprocessing methods. Details can be found in the code available as part of the Supporting Info.

In the second approach, we trained a logistic regression classifier directly on the training set without any access to the validation data, to assess performance under low-data constraints. In both cases, the test set was evaluated only once to ensure an unbiased assessment of the performance. 
All classification models and optimization routines were implemented using the \textit{scikit-learn} library in Python (version 1.7).\cite{scikit-learn}

\textbf{5. Object Detection with YOLOv11}
YOLO11 object detection model, was fine-tuned on the three datasets using the Ultralytics implementation.\cite{Khanam2024YOLOv11} We applied YOLO11-L (large) for cubes and dots datasets and YOLO11-X  (extra large) for the triangles dataset. Training was conducted using bounding box annotations in YOLO format. The models were initialized with pretrained weights of the corresponding model architectures. Optimization was performed using stochastic gradient descent with YOLO’s default learning rate of 0.01. Due to the small size of the datasets, we did not specify a batch size, and the framework automatically adjusted it based on the number of available training images. Training was conducted over 70 epochs for cubes and dots and 50 epochs for triangles data. We applied the default YOLO’s built-in data augmentation. The pipeline comprised random scaling, translation, flipping, and color transformations.

All the code and the datasets can be found in the following repository: https://github.com/freida20git/nanoparticle-classification/  

\textbf{6. Zero-shot nano-particle recognition using ChatGPT o4-mini-high}
For a layman the natural first stop is a 'zero-shot' model in which no training or finetuning is involved. To this end ChatGPT, was employed. using 'model o4-mini-high'.  This platform in fact includes a native visual language model (VLM) - a deep-learning pipeline comprised of convolutional layers together with transformer-based attention layers - that segments  images and runs a learned shape-classification head.\cite{bordes2024vlm}

In fact ChatGPT was run only for classification but not for segmentation. For segmentation ChatGPT immediately turned to classical image processing techniques, such as edge-detection, Otsu-thresholding\cite{otsu1979threshold} and watershed segmentation, requiring interactive tuning of one or more parameters to obtain even passable results. This is far from the goal of a zero-shot method easily accessible to the nanoparticle professional, unitiated in image processing techniques.

For classification , a minimal amount of context was first provided - the total image along with the information that it displays a SEM of silver nanoparticle.  Thumbnails of particles cropped from the main image  were then uploaded  and the LLM was instructed to classify each images as 'triangle', 'nearly triangular' or 'non-triangular'.   (See Figure~\ref{fig:binary_masks_thumbnails} ahead in the Results section).
In light of the preceding comments regarding segmentation, the image crops used, were those produced by SAM - specifically, the binary masks of the particles were input for classification, as noted above for DINOv2-based classification. Thus the comparison to ChatGPT specifically benchmarks the DINOv2 feature extraction + light classification head but not the segmentation stage.
The results are described in the next section.

\textbf{7. SEM Microscopy Imaging}

The size and shape of the NPs were investigated using an environmental scanning electron microscope (E-SEM) (Quanta FEG 250, FEI) with high vacuum. Particle size measurement was carried out manually using Image-J software. (For example see Figure~\ref{fig:manual-characterization} in the Appendix.)

\section{Results}
\begin{figure}[H]
\centering
\includegraphics[width=1\textwidth]{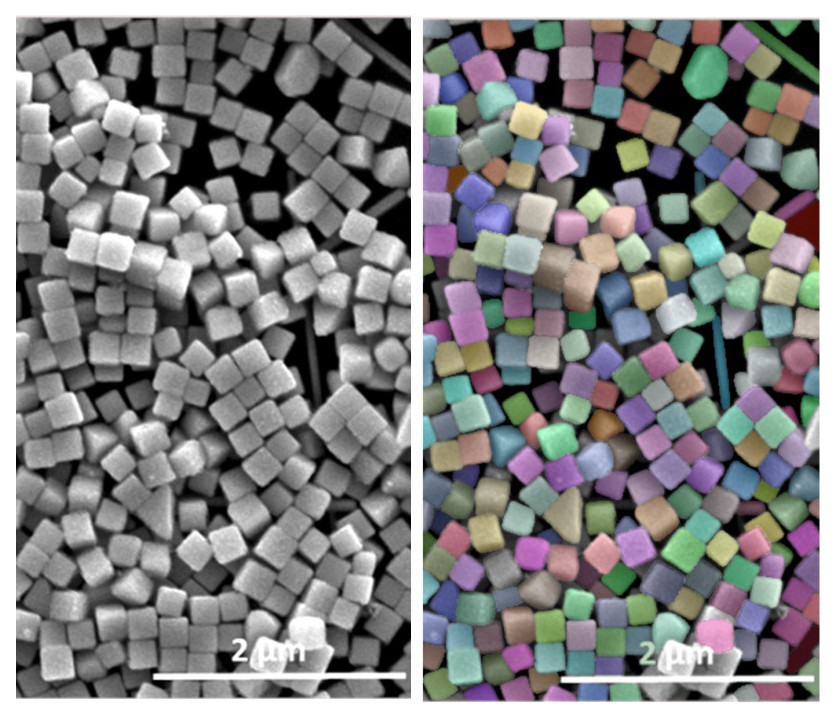}
\caption{Output of the Segment Anything Model (SAM) applied to a Scanning Electron Microscopy (SEM) image of cubes nanoparticles dataset. Each segmented particle is overlaid with a distinct color, representing a separate segmentation mask automatically generated by SAM. These individual masks were used to crop the original image into isolated particle instances for subsequent shape classification.  }
\label{fig:main SAM ilustration}
\end{figure}

To disassemble the SEM images into individual nanoparticles, the Segment Anything Model (SAM) was employed, which automatically generated segmentation masks for each particle. 
Figure~\ref{fig:main SAM ilustration} illustrates the SAM-based segmentation applied to the mixture of cubic and pyramid-shaped nanoparticles dataset. 
As shown, SAM effectively segmented not only the clearly visible particles but also those that are partially occluded or located at the image boundaries. The success in segmenting a multitude of densely packed objects is particularly noteworthy. The segmentation of overlapping images is an outstanding problem in image processing which proves challenging even for state-of-the-art object detection algorithm such as YOLO. As described shortly, the latter can be employed, yet a significant effort must be invested in training. Moreover, the segmented objects are identified by a bounding box alone, rather than a precise segmentation of the object specified by a binary mask, known as 'instance segmentation'. As exemplified by the distinct colorized regions, SAM provides nearly perfect 'tightly fitting' instance segmentation. No training is required - operation is 'zero-shot'. 

Images of densely packed, overlapping objects are common among  nanoparticle SEM results. Hence the results embodied by the figure, of themselves signify a major advance in automated analysis of nanoparticle SEM imaging. Moreover, the expertise and experience of researchers involved in nanoparticle synthesis - and imaging - typically lie elsewhere than  (neural-) network training; the simplicity in applying SAM thus heralds its widespread adoption by specialists in the field of nanoparticle synthesis.

\begin{figure}
    \centering
    \includegraphics[width=0.5\linewidth]{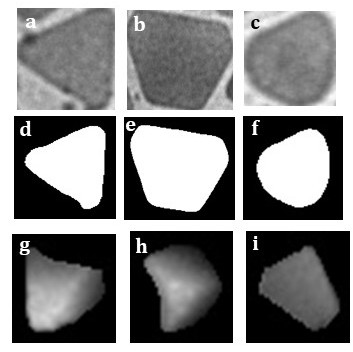}
    \caption{Binary masks produced by SAM1 and cropped images for triangular particles These were used for training and testing in the classification stage. \textbf{a,b,c }: Triangle, Nearly-triangular, Non-triangular samples, respectively. \textbf{d,e,f} : Corresponding masks.\textbf{g,h,i} : Cropped images of a pyramid  and two cubes respectively. We utilized the mask to isolates the object by removing the irrelevant background regions, ensuring that DINOv2 embeddings are computed solely from the segmented structure of interest.}
    \label{fig:binary_masks_thumbnails}
\end{figure}

Next, from the masks produced by SAM, bounding boxes enclosing each segmented shape were derived. These bounding boxes were used to crop the original image into isolated particle instances, enabling shape-specific feature extraction and classification in the next stages of our pipeline. A few examples of cropped images of single nanoparticles are displayed in Figure~\ref{fig:binary_masks_thumbnails}a-c, together with the corresponding binary masks in Figure~\ref{fig:binary_masks_thumbnails}d-f.

\begin{figure}[H]
\centering
\includegraphics[width=1.1\textwidth]{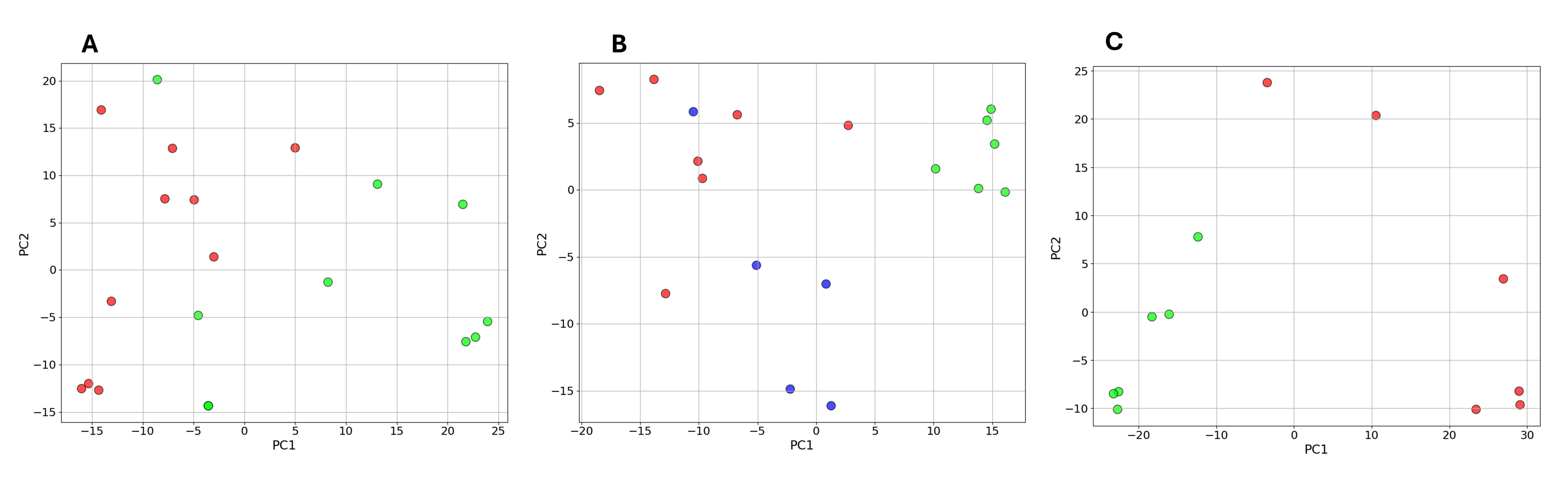}
\caption{PCA projection of DINOv2 feature embeddings from the training set. Each point represents a sample in the reduced 2D space, capturing the principal variance in the high-dimensional embedding space. Color coding indicates class labels. A. Green: cubes, Red: pyramids. B. Green: circles, Red: triangles, blue: truncated triangles C. green: dots, green non dots. It can be observed that the training dataset is very modest in size and the points form well-defined clusters, highlighting the strong representational capacity of DINOv2 embeddings.}
\label{fig:Main_PCA}
\end{figure}

Each cropped particle image was passed through the pretrained DINOv2 model to obtain a feature vector encoding its visual properties. These vectors were then subjected to Principal Component Analysis (PCA), a dimensionality reduction technique, to visualize their distribution in two dimensions. The resulting 2D projection, depicted in Figure~\ref{fig:Main_PCA}, showed generally clear and well-separated clusters corresponding to the different particle types, suggesting that DINOv2 effectively captures meaningful and discriminative features.  As noted in the previous section (see Methods)  superior results were obtained by imputing the binary masks produced by SAM directly to the classifier.

\begin{table}[H]
\centering
\caption{Performance comparison across different methods}
\label{tab:main_results}
\begin{tabular}{|c|l|l|r|r|r|}
\hline
\hline
\textbf{Dataset} & \textbf{Class} & \textbf{Method\textsuperscript{*}
}& \textbf{Recall} & \textbf{Precision} & \textbf{F1} \\
\hline
1 & Cubes     & DINOv2     & 0.98 & 0.99 & 0.99 \\
  &           & DINOv2 LR  & 0.95& 0.99 & 0.97\\
  &           & YOLO       & 0.59 & 0.94 & 0.73 \\
 & & ChatGPT o4-mini-high& N/A& N/A&N/A\\
 & & & & &\\
  & Pyramids  & DINOv2     & 0.83 & 0.77 & 0.80 \\
  &           & DINOv2 LR  & 0.83& 0.56& 0.67\\
  &           & YOLO       & 1.00 & 0.10 & 0.12 \\
 & & ChatGPT o4-mini-high& N/A& N/A&N/A\\
\hline
2 & Triangle           & DINOv2     & 0.99& 0.98& 0.98\\
  &                    & DINOv2 LR  & 0.98 & 1.00 & 0.99 \\
  &                    & YOLO       & 0.91 & 0.52 & 0.66 \\
 & & ChatGPT o4-mini-high& 0.69
& 0.98
&0.81
\\
   &           &        &  &  &  \\
  & Trunc. Triangles   & DINOv2     & 1.00& 0.85& 0.92\\
  &                    & DINOv2 LR  & 0.91 & 0.71 & 0.80 \\
  &                    & YOLO       & 0.21 & 0.82 & 0.33 \\
 & & ChatGPT o4-mini-high& 0.18
& 0.09
&0.12
\\
   &           &        &  &  &  \\
  & Circles            & DINOv2     & 0.73 & 1.00& 0.84\\
  &                    & DINOv2 LR  & 0.73 & 0.80 & 0.76 \\
  &                    & YOLO       & 0.65 & 0.82 & 0.73 \\
 & & ChatGPT o4-mini-high& 1.00
& 0.44&0.61
\\
\hline
3 & Dots      & DINOv2     & 0.99& 1.00 & 1.00\\
  &           & DINOv2 LR  & 0.99 & 1.00 & 1.00 \\
  &           & YOLO       & 0.1 & 0.78 & 0.17 \\
   &           &        ChatGPT o4-mini-high&  N/A&  N/A&  N/A\\
 & & & & &\\
  & Non-dots  & DINOv2     & 1.00 & 0.88 & 0.93\\
  &           & DINOv2 LR  & 1.00 & 0.88 & 0.93\\
  &           & YOLO       & 0.57 & 0.94 & 0.71 \\
 & & ChatGPT o4-mini-high& N/A& N/A&N/A\\
 & & & & &\\
\hline
\multicolumn{2}{|c|}{} & \textbf{DINOv2}     & \textbf{0.93}& \textbf{0.92}& \textbf{0.92}\\
\multicolumn{2}{|c|}{\textbf{Average}} & \textbf{DINOv2 LR}  & \textbf{0.91}& \textbf{0.85}& \textbf{0.87}\\
\multicolumn{2}{|c|}{} & \textbf{YOLO}       & \textbf{0.58}& \textbf{0.70}& \textbf{0.49}\\
 & & \textbf{ChatGPT o4-mini-high} (triangles image)& 0.62& 0.50&0.51\\
\hline

\end{tabular}
\setcounter{table}{1}
\caption{\textsuperscript{*}{"DINOv2" refers to results obtained after hyperparameter optimization on the validation set. "DINOv2 LR" denotes a logistic regression classifier trained on DINOv2 features using only the training set, without access to the validation set. "YOLO" corresponds to predictions made by the YOLOv11 object detection model "N/A" =  not applicable: ChatGPT was not applied to the Dots or Cubes.}}
\end{table}

\begin{figure}[H]
\centering
\includegraphics[width=1\textwidth]{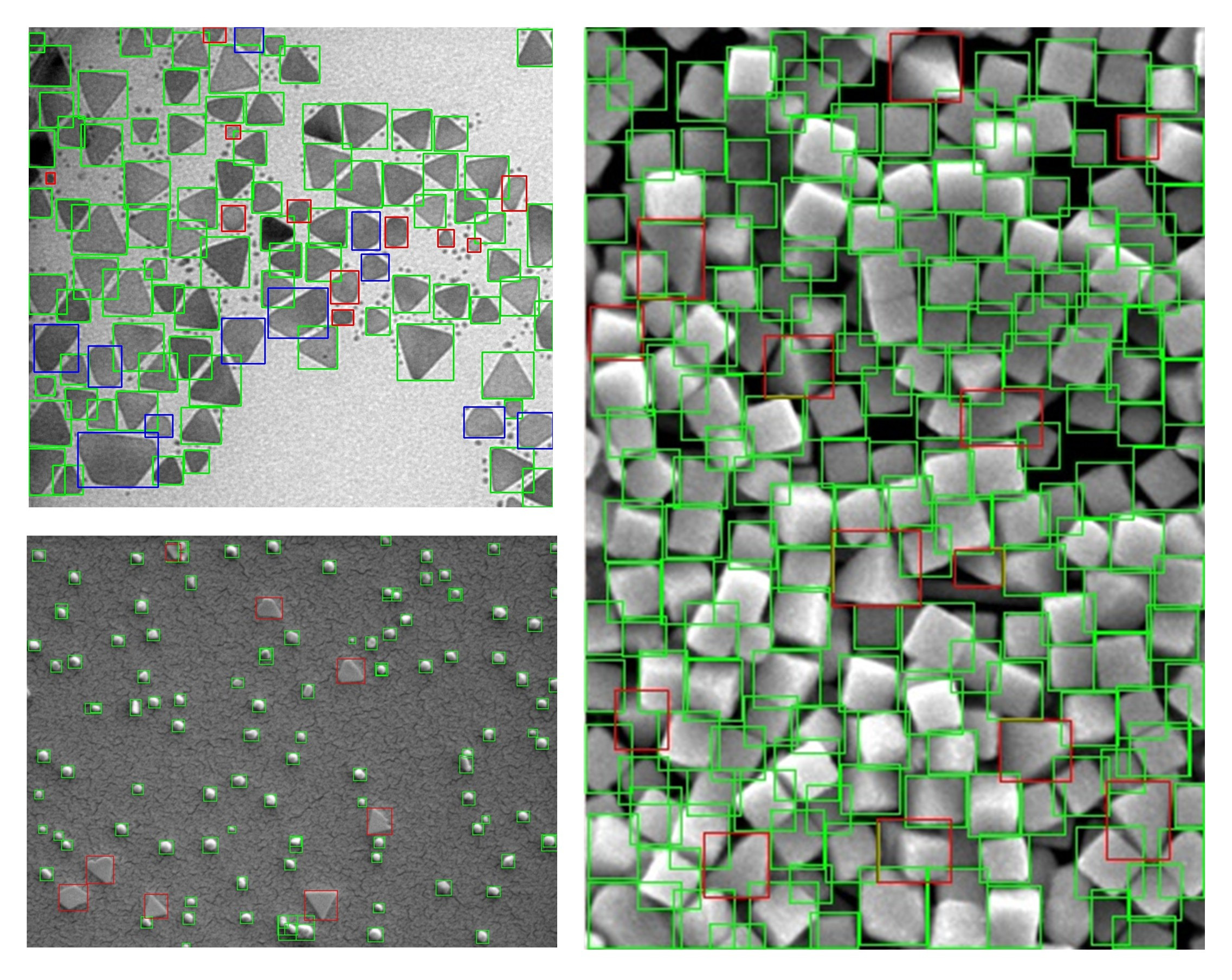}
\caption{Detection and classification results of nanoparticles. Particle regions were segmented using the Segment Anything Model (SAM), followed by classification using DINOv2-based feature extraction. Each segmented particle is enclosed in a bounding box coloured by its predicted class. predicted class: image A: green for circles, blue for triangles, and red for truncated triangles. image B: blue for cubes and green for pyramides}
\label{fig:nanoparticle_bbox}
\end{figure}

Table 2 presents the performance metrics of the different classifiers applied to the nanoparticle datasets. The DINO-based approach was evaluated in two methods: “DINOv2” refers to results obtained after hyperparameter optimization on the validation set, while “DINOv2 LR” denotes a logistic regression classifier trained on DINOv2 features using only the training set, without using the validation set for hyperparameters optimization. 
Despite the small training set and significant class imbalance the classifiers achieved high precision and recall across all the categories even at the challenging minority-classes.
Notably, DINOv2 LR achieved a high average F1 performance (0.87) close to the optimized DINOv2 classifier (0.92), despite being trained without any classifier optimization. Though surprising, this in effect corroborates the optimal nature of DINOv2's feature extraction.

As described in the previous section the first baseline for comparison to the achievements of  SAM1+DINOv2 was taken to be ChatGPT, model o4-mini-high, running its internal vision pipeline, 'zero-shot' - i.e. with no fine-tuning on training data. ChatGPT was run for the triangles alone, and only for classification - using the binary masks output by the  SAM1 segmentation as input, just as for the SAM-DINO pipeline. Insofar as real grey-scale image crops introduce uncertainty and noise the results really express an upper bound on the performance that can be expected with this method - say if ChatGPT performed both segmentation and classification.

The performance on the whole was fairly mediocre, as expressed by the data in Table 2. It lagged far behind that of DINOv2. and somewhat below that of YOLO as well, discussed next. Moreover it was less convenient than one might expect as elaborated ahead (see Discussion).

To further evaluate the performance of our SAM+DINOv2 pipeline, we compared it to a traditional deep learning approach using YOLOv11 for direct nanoparticle classification. The comparison results are presented in Table 2.  Across all datasets and classes, except for the circles detection in the triangles dataset both DINOv2-based approaches consistently outperformed YOLO, achieving significantly higher recall, precision, and F1 scores.  In contrast, YOLO struggled especially on minority or irregular classes, such as pyramids and truncated triangles, with F1 scores often falling below 0.4. Examining YOLO's bounding box predictions reveals that, while the model generally successfully detected the presence of nanoparticles, it failed to generalize across shape classes—frequently misclassifying visually similar particles. 

Figure~\ref{fig:nanoparticle_bbox} illustrates the DINOv2 predictions from the test SEM images, showcasing the three different datasets. In each image the corresponding nanoparticle shapes classes listed in Table 2 can be visually examined along with the boundary boxes of the prediction.
Notably, within the triangles image, the classifier was able to further differentiate subtle variations in shape, successfully separating regular triangular particles from their truncated counterparts—highlighting the model’s sensitivity to fine-grained differences in morphology. In the cubes nanoparticle image, despite the presence of overlapping and partially occluded particles, the classifier successfully distinguished between the shape categories. Although the dataset was highly imbalanced, with a dominant presence of cubic nanoparticles, the classifier was still able to detect and correctly classify the minority class of contaminating pyramid-shaped particles. This highlights both  robustness in the face of limited training data as well as sensitivity to rare morphological variations.

\section{Discussion}

The results displayed in the PCA plots are noteworthy for their simplicity. The embeddings created by DINOv2  are highly non-linear functions of the input and very high-dimensional. One might not expect that PCA,  based on a linear projection, would render well-defined and well-separated clusters in low-dimensions. More often a more complex nonlinear approach such as t-SNE is necessary for  visualization.\cite{maaten2008tsne} The fact that PCA is sufficient can be viewed as affirmation of the quality and power of the DINOv2 features extraction in describing nanoparticle morphologies.

The PCA  representation displayed in Figure~\ref{fig:Main_PCA} in fact conveys  physical information which potentially can be of great use to the scientist performing the chemical synthesis. Frequently the nanoparticles undergo a process of 'differentiation'. During the synthesis process, the nanocrystals start their development with indistinct, ill-defined morphologies. As they grow, their morphologies crystallize into well-defined shapes, readily separated (visually) into a small number of distinct categories. Spatially it is not uncommon for the nanocrystals to develop initially as 'buds' along a main stem; these break off as the process matures, and the particles continue development. Conversely, in another common scenario, nanocrystals develop  from well (spatially) separated seeds in a solution and undergo aggregation in late stages of the process. The formation of conglomerate morphologies such as multi-petaled 'florettes' is a familiar result.
Qualitatively, the PCA plots - and in particular the degree of separation between different morphological classes - can  offer valuable insight into the stage of development. 

Quantitatively the inter-class ('between-class') variance of the clusters can be used as a metric. For instance in the first scenario described above, initially the (proto-) crystals do not have definite form, and so comprise a single cluster, albeit spread-out in feature space. The inter-class variance is thus zero.  As the crystals differentiate into well-defined shapes - say cubes and pyramids, for instance - this inital cluster splits into a few smaller clusters which progressively separate from one another. The inter-class variance thus increases with time.

The intra-class ('internal') variance also provides quantitative information. As just noted, the initial cluster will display a PCA-plot  with a large variance. The smaller clusters which emerge represent well-defined shapes and will thus be more concentrated. The  total intra-class variance will thus be lower. This trend will continue, as within each group the morphologies converge. The intra-class variance will thus decrease as the reaction progresses.

The separation between the clusters can also quantified, for instance using the 'Silhouette Score'. \cite{rousseeuw1987silhouettes}
A schematic plot illustrating the evolution of the inter- and intra-class variance, as will as the silhouette score can be seen in Figure\ref{fig:Siluate_example} in the appendix.

A few technical comments on the use of ChatGPT are in order:  When a single image is uploaded and qualitative queries are made by default  the model invokes its native 'vision pipeline'. However in many cases the model responds by creating a flow comprised of classical image processing methods;  these are executed by  generating  code in Python - using routines from OpenCV, Sci-kit Image packages -  which is run internally.\cite{bradski2000opencv,scikit-learn}  Such pipelines as a rule involve parameters, and rely on the user for interactive calibration. The results are often less accurate than the internal vision pipeline (when this can be invoked).
This is typically the scenario when quantitative queries are made - how many particles, etc. It is currently possible - using a 'plus' subscription - to upload up to 10 image files a time. For a test set containing 100+ images, this is inconvenient at best; worse it runs the risk of a a system crash - 'kernel error' - in the middle of the analysis. The common alternative is to upload all the files in a single .zip file;  this approach however does not allow analysis with the internal vision pipeline and appears amenable only to treatment via a classical image processing pipeline. Presumably these difficulties may be avoided or resolved using the 'Professional' subscription, or an API license. But the disparities in terms of budget and ease-of-use constitute a definite advantage for the SAM2 /DINOv2-based approach.

Following segmentation and classification the next step naturally is to perform quantitative analyses of the particles. Currently following segmentation and classification it is a simple matter to report the exact number of each nanoparticle type. Beyond this, the size of the particles is always of interest. For the triangles this means a determination of the lengths of the edges. In two dimensions this follows directly from determination of the positions of the vertices. However for the nanocubes, considering the image  in Figure~\ref{fig:nanoparticles_training}, the three-dimensional nature of the cubes and their arrangement in space means  a depth-map is necessary to measuring the edge lengths.  Assuming an orthographic projection, this can be inferred from the two-dimensional images using a foundation model such as Depth Anything.\cite{depth_anything_v2} Initial results with this approach are promising (see Amitay, Mandelbaum, \textit{et al}. work-in-progress). For another class of nanoparticle,  the core-shell constructions - not considered here - the determination of the radius (ie. thickness) of each layer is important. 

For the triangles the values of the angles of each particle is not fixed but varies from speciment to specimen.  Determination of the angles again follows directly from knowledge of the positions of the vertices.

\begin{figure}[H]
    \centering    
    \includegraphics[width=0.5\linewidth]{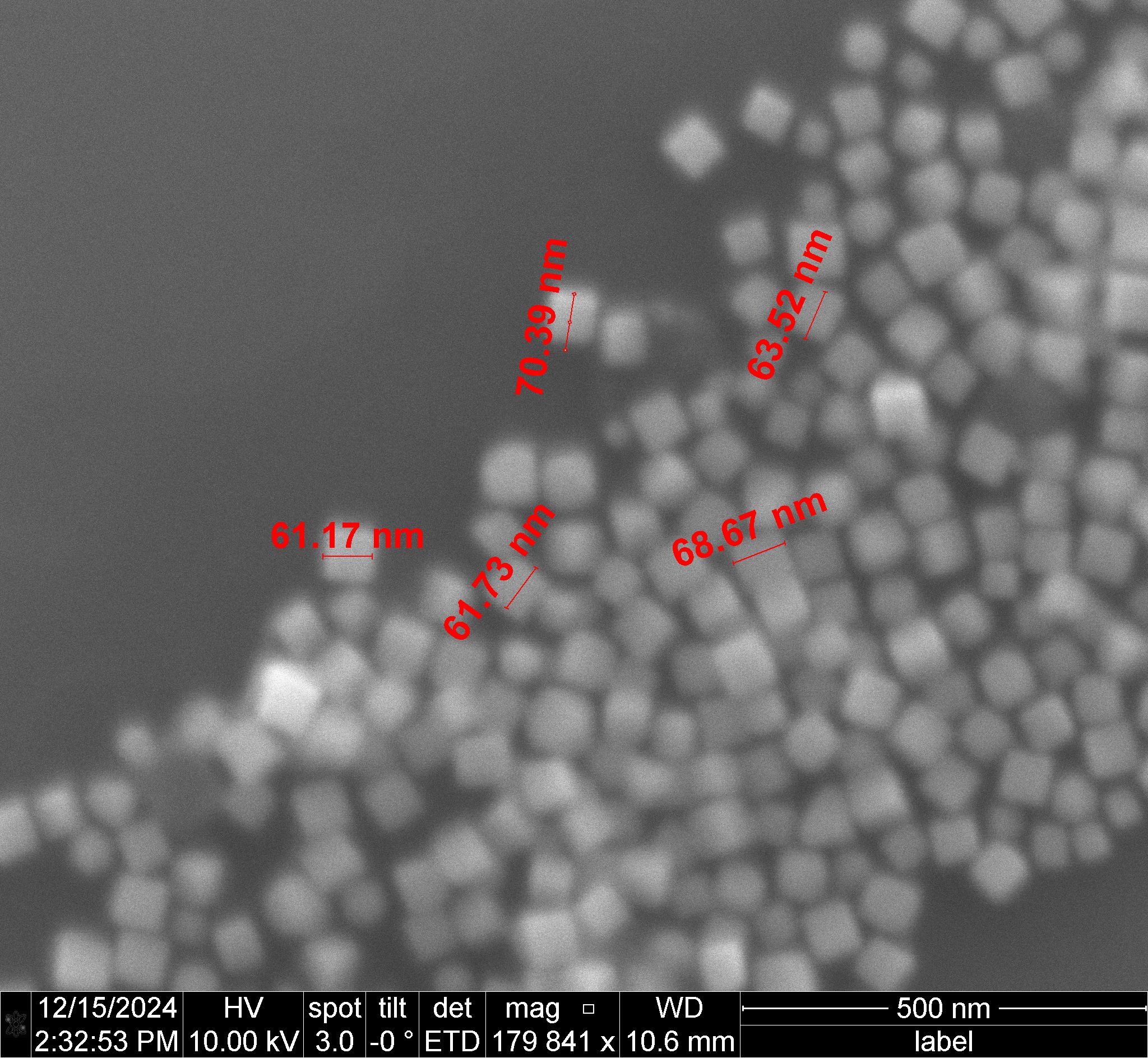}
    \caption{Currently, nanoparticle characterization requires (semi-)manual measurements, as demonstrated here. For batches the size of this nano-cube batch, this process can be very onerous. Blurring is evident, originating in the settings of the (SEM) imaging device. }
    \label{fig:manual-characterization}
\end{figure}

The quality of nanoparticles produced in a batch is of course also a fundamental characteristic. For the triangles this means quantifying the 'sharpness'  - the degree of angularity vs. roundness  of the vertices. The sharpness of the corners also serves to gauge the quality of the cubes. However, this can be challenging to determine when the image contains blurring due to imperfect focus, as for instance in Figure~\ref{fig:manual-characterization}.

Determination of the distribution of orientations in which the particles lie can be of importance for the triangles and the cubes and in fact for any class of nanoparticle which are not circular. For instance for fluorescent nanorods the distribution determines the degree of polarization of emitted radiation.\cite{amit2011semiconductor}

For all such quantitative measurements - orientation, angular composition, 'sharpness', and size - the goal is to make an exhaustive analysis and assemble the complete distribution for all the particles in the image.

The results reported here demonstrate that  the combination of SAM segmentation and DINOv2 embeddings provides robust and discriminative features, enabling accurated detection and shape-based classification of nanoparticle SEM images. 
In the process this highlights the potential in general for applying foundation models such as SAM and DINOv2 to high-resolution scientific imaging tasks beyond natural image domains. Compared to traditional deep learning approaches, this methodology offers significant advantages. First, it achieves superior classification accuracy, particularly in challenging microscopy scenarios involving limited training data. Second, it is much more accessible and resource-efficient: it eliminates the need for large annotated datasets, GPU-hardware and or fine-tuning complex deep learning networks. This zero-shot, lightweight pipeline enables easier and effective analysis, making it a better solution for scientific research environments.  

Working with  SAM and DINOv2 may be further streamlined by using a large-language model (LLM)  as a front-end. For instance ChatGPT smay be used as generate scripts (and Jupyter notebooks) in Python to apply foundation models.  Currently the LLM will not run SAM or DINO directly, and a few iterations - in which run-time error messages are fed back to the LLM and an improved script is obtained - are typically necessary.  However continual advances in LLM capabilities are expected to facilitiate direct communication with auxiliary programs and devices as well, particularly employing the recently released MCP protocol.\cite{bloom2025mcp}

Thus in future work, the methodology embodied by the SAM1 + DINOv2 pipeline could be extended into fully automated analysis pipelines that incorporate batch processing, real-time feedback during synthesis, or even integration with active learning loops for iterative improvement—bringing robust nanoparticle characterization closer to real-time and hands-free operation.
As foundation models continue to advance, their ability to generalize across non-natural imaging domains such as scanning electron microscopy (SEM) is expected to improve, offering tools which are both increasingly powerful and more easily applied for automated nanoparticle detection and shape classification.

\section{Conclusion}
In this study, a zero-shot, foundation model methodology was utilized for the classification of nanoparticles in SEM images. By combining the Segment Anything Model (SAM) for segmentation with DINOv2 for feature extraction and downstream classification, it was demonstrated that complex particle morphologies can be accurately detected and categorized without the need for task-specific training, extensive labeling, or GPU resources. This approach provides a general and scalable framework for analyzing nanoparticle shapes across varied SEM imaging conditions, offering a significant improvement in both usability and robustness over conventional deep learning pipelines. Along the way the extracted features were seen to convey valuable information about the progress of the chemical synthesis.
As foundation models continue to evolve and become more attuned to scientific domains, one anticipates that such zero-shot methods will only become more accurate, adaptable, and widely applicable in nanomaterials research and other fields reliant on microscopy image analysis.

\bibliography{mybib}
\bibliographystyle{unsrt}

\clearpage
\section{Appendix}

\begin{figure}[H]
    \centering
    \includegraphics[width=0.5\linewidth]{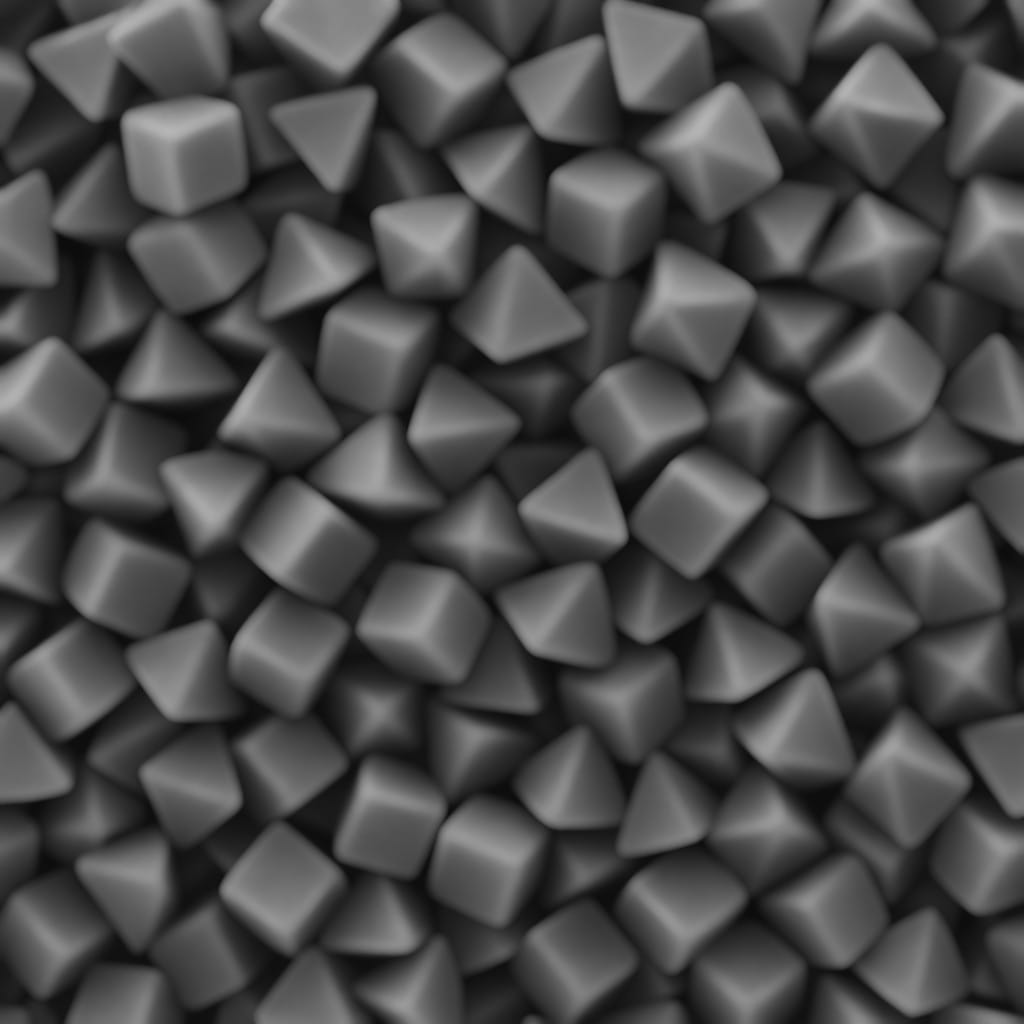}
    \caption{The synthetically generated SEM image used for the validation set.}
    \label{fig:synthetic_generated_validation}
\end{figure}

\begin{figure}[ht]
    \centering
    \includegraphics[width=0.8\linewidth]{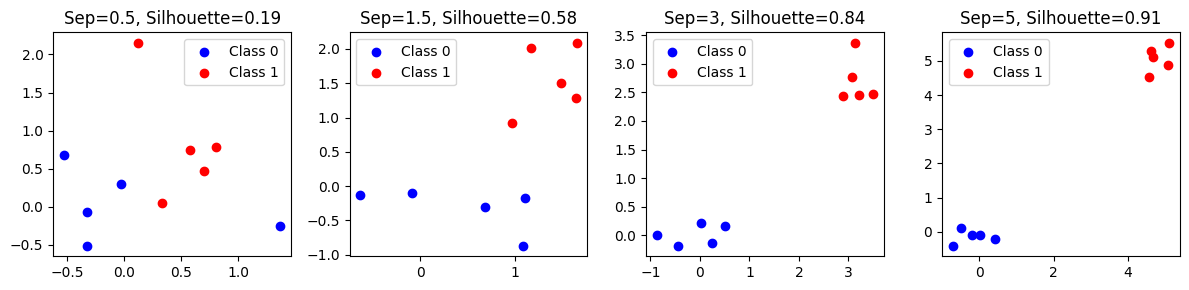}
    \caption{Clustering Metrics: from left to right the clusters separate and converge. The \textbf{inter-class variance} increases, and with it the \textbf{Silhouette score}, while the \textbf{intra-class variance} decreases.}
    \label{fig:Siluate_example}
\end{figure}

\textbf{Nanoparticle Synthesis}

\textbf{Synthesis of Ag (nanocubes)} 
Reprinted with permission from\cite{huri2024spr}

\textbf{Ag-NPs Synthesis (triangles)}
Typically, an aqueous solution of silver nitrate (STREM 99.9\%, 0.1 mM in Deionized Water (DW), 25 mL), trisodium citrate (30 mM, 1.5 mL), Polyvinyl pyrrolidone (PVP, Aldrich, Mw ~ 29000 g mol ,0.7 mM, 1.5 mL), and hydrogen peroxide (30 wt.\%, 60 µL) were combined and vigorously stirred at room temperature in the presence of air. To this mixture, sodium borohydride (100 mM, 100-150 µL) was rapidly injected, generating a colloid that was pale yellow in color. After 30 min, the colloid darkened to a deep-yellow color, indicating the formation of small Ag-NPs (as evidenced by UV-Vis spectroscopy and the identification of the surface plasmon resonance). Over the next several seconds, the color of the colloid continued to change from yellow to red.

\end{document}